\documentclass{svproc}
\usepackage{xspace}

\usepackage{amsmath}
\usepackage{amssymb}
\usepackage{algorithm}
\usepackage{algpseudocode} % provides \Require, \Ensure, \State, etc.
\usepackage{booktabs}

\usepackage[colorlinks=true,linkcolor=blue,citecolor=blue,urlcolor=blue]{hyperref} % if you use hyperref
\usepackage[nameinlink,capitalise]{cleveref} % load after hyperref
\usepackage{url}

\usepackage{multirow}
\usepackage{ulem}
\usepackage{pifont}
\usepackage{soul}
\usepackage{graphicx}

\usepackage{enumitem}
\usepackage{gensymb}

\usepackage{subcaption}

\usepackage{colortbl}
\usepackage{makecell}
\usepackage{rotating}
\usepackage{arydshln}

\newcommand*{\eg}{e.g.\@\xspace}
\newcommand*{\ie}{\textit{i.e.}\@\xspace}

\newcommand*{\cf}{\textit{cf.}\@\xspace}

\newcommand{\name}{SF3D-RGB}%

\begin{document}
\mainmatter              % start of a contribution
\title{\name{}: Scene Flow Estimation from Monocular Camera and Sparse LiDAR}
\titlerunning{\name{}}  % abbreviated title (for running head)

\author{Rajai Alhimdiat\inst{1}\thanks{These authors contributed equally.} \and Ramy Battrawy\inst{2}\footnotemark[1] \and Ren{\'e} Schuster\inst{2} \and
Didier Stricker\inst{2} \and Wesam Ashour\inst{1}}
\authorrunning{Rajai Alhimdiat et al.} % abbreviated author list (for running head)
\tocauthor{Rajai Alhimdiat, Ramy Battrawy, Didier Stricker and Wesam Ashour}
\institute{Islamic University of Gaza, Gaza, Palestine,\\
\email{rhimdiat@students.iugaza.edu.ps, washour@iugaza.edu.ps}
\and
German Research Center for Artificial Intelligence (DFKI),\\
Kaiserslautern, Germany \\
\email{\{ramy.battrawy,rene.schuster,didier.stricker\}@dfki.de}\\
}

\maketitle              % typeset the title of the contribution
% \footnotetext[*]{These authors contributed equally.}

\begin{abstract}
Scene flow estimation is an extremely important task in computer vision to support the perception of dynamic changes in the scene. For robust scene flow, learning-based approaches have recently achieved impressive results using either image-based or LiDAR-based modalities. However, these methods have tended to focus on the use of a single modality. To tackle these problems, we present a deep learning architecture, \name{}, that enables sparse scene flow estimation using 2D monocular images and 3D point clouds (\eg, acquired by LiDAR) as inputs. Our architecture is an end-to-end model that first encodes information from each modality into features and fuses them together. Then, the fused features enhance a graph matching module for better and more robust mapping matrix computation to generate an initial scene flow. Finally, a residual scene flow module further refines the initial scene flow. Our model is designed to strike a balance between accuracy and efficiency. Furthermore, experiments show that our proposed method outperforms single-modality methods and achieves better scene flow accuracy on real-world datasets while using fewer parameters compared to other state-of-the-art methods with fusion.
% We would like to encourage you to list your keywords within
% the abstract section using the \keywords{...} command.
\end{abstract}

\section{Introduction} \label{Introduction}
Scene flow estimation aims to perceive the 3D motion field of a dynamic scene. This information is crucial for a variety of scene understanding tasks, such as in robotics, autonomous driving, and augmented reality. Recently, learning-based approaches have achieved impressive results in scene flow estimation using single modalities such as images (\eg, stereo systems) or LiDAR measurements.

Image-based solutions tend to estimate scene flow by constructing a high-dimensional cost volume, which is then refined and decoded into a dense scene flow \cite{aleotti2020learning,ilg2018occlusions,jiang2019sense,saxena2019pwoc}. However, this density comes at the cost of efficiency.
Furthermore, the accuracy of these methods is entirely dependent on the quality of the images, resulting in inaccurate scene flow if the images contain textureless areas.

LiDAR sensors are less sensitive to environmental conditions and can provide more accurate 3D measurements than image-based systems. 
This opens a wide door for many approaches that take advantage of LiDAR to estimate scene flow \cite{behl2019pointflownet,gu2022rcp,kittenplon2021flowstep3d,liu2019flownet3d,puy20flot,wei2021pv}.
However, LiDAR sensors acquire unstructured data and processing it is much more difficult, requiring either intermediate representations (\eg, Lattice \cite{gu2019hplflownet}) at the expense of resolution or special processing to define local regions using k-nearest neighbors (\textit{k}-NN) search, the latter being more accurate but less efficient in terms of runtime. In addition, LiDAR-based solutions have difficulty matching regions with homogeneous geometry or coplanar regions. 

The fusion of both modalities (\ie, LiDAR and images) has recently become an important area of research, where LiDAR provides accurate 3D measurements, and 2D images provide rich texture information. Both can be used together to overcome the limitations of each modality. The result is a more robust 3D scene flow estimation.

In this context, some scene flow solutions use unidirectional fusion with LiDAR represented as disparity/depth and image data, but both are based on 2D representations, such as LiDAR-Flow \cite{battrawy2019lidar} and DeepLiDARFlow \cite{rishav2020deeplidarflow}, the latter is a learning-based approach that estimates scene flow at multiple scales. More recently, RAFT-3D \cite{teed2021raft} develops a geometric consistency neural network along with a rigidity constraint and iteratively refines the optical flow with the SE3 motion field of the rigid objects. However, all of the above dense scene flow estimators are far from real-time use and require high memory and processing due to the construction of cost volume (\ie, the construction of correlation candidates).

Conversely, other approaches \cite{wang2020flownet3d++,wang2021festa} exploit the 3D structure of LiDAR point clouds by directly fusing spatial locations (\ie, xyz coordinates) with corresponding RGB intensities through early fusion. 
Both 2D and 3D fusion strategies tend to lose modality-specific robustness.
For example, projecting a point cloud onto 2D can map multiple points to the same pixel, causing loss of geometric detail. Moreover, applying image-domain convolutions becomes problematic because LiDAR data is sparse and low-resolution, unlike dense RGB images.
On the other hand, bringing RGB features into the 3D domain can diminish their inherent density, since they must be aligned with the sparse LiDAR structure.

% However, these methods rely on complex architectures with a large number of parameters, which makes them less efficient.

To address these issues, we introduce \name{}, an end-to-end framework that predicts sparse scene flow while achieving a strong balance between accuracy and efficiency. 
Our method leverages the strengths of both modalities by extracting RGB features in the 2D image domain and fusing them with corresponding LiDAR features in the 3D domain, enabling effective sparse 3D scene flow estimation with high accuracy. 
Our LiDAR feature extraction module computes 3D point cloud features using the backbone of PointNet \cite{qi2017pointnet} and hierarchically extracts image features through a Feature Pyramid Network (FPN) \cite{lin2017feature}. 
In our fusion module, we combine coarse-scale image features with the corresponding 3D point cloud features, yielding more distinctive and robust point cloud representations.
We then pass the fused features to a graph matching module based on optimal transport (Sinkhorn algorithm) \cite{cuturi2013sinkhorn,peyre2019computational}, followed by a refinement stage, both inspired by FLOT \cite{puy20flot}. 
Unlike FLOT, our method achieves more reliable correlations thanks to the stronger representations produced through RGB–LiDAR fusion.

Our solution is more efficient than dense scene flow methods \cite{rishav2020deeplidarflow,teed2021raft} and achieves higher accuracy than early fusion methods such as \cite{wang2021festa,wang2020flownet3d++}.\\
We summarize the main contributions as follows:
\begin{itemize}[noitemsep,nolistsep,topsep=1pt,leftmargin=*,label=\textbullet]
\item We propose \name{}—an efficient end-to-end neural network architecture for sparse scene flow estimation.
\item Our architecture fuses LiDAR point cloud features with monocular RGB features to robustly compute the optimal assignment matrix from their correlations.
\item We design our architecture to be lightweight, using few parameters while offering a strong balance between accuracy and efficiency.
\item Our method surpasses early fusion approaches in accuracy and outperforms other state-of-the-art methods in efficiency on the FlyingThings3D (FT3D) benchmark \cite{mayer2016large}. We further evaluate scene flow accuracy on the real-world KITTI dataset \cite{geiger2012we,menze2015object}, both without fine-tuning and with fine-tuning.
\end{itemize}

\section{Related Work} \label{Related Work}
\subsection{An Individual Modality for Scene Flow Estimation}
Many deep learning methods estimate scene flow directly in the image domain by using image-based modalities: monocular cameras \cite{brickwedde2019mono,chen2019pct,luo2019every,yang2020robust}, stereo systems \cite{aleotti2020learning,ilg2018occlusions,jiang2019sense,saxena2019pwoc}, or RGB-D cameras \cite{herbst2013rgb,jaimez2015motion,jaimez2015primal,quiroga2014dense,yoshida2017time,qiao2018sf}. 
However, all image-based modalities are highly dependent on image quality and can suffer from noisy geometry, such as in areas with poor lighting, shadows, and reflective objects. 

Nowadays, LiDAR-based scene flow estimation is gaining more attention due to its accurate measurements. In this context, many works use advanced neural architectures for scene flow \cite{behl2019pointflownet,kittenplon2021flowstep3d,gu2022rcp,liu2019flownet3d,puy20flot,wei2021pv}. Some of these methods are designed to estimate scene flow at multiple scales \cite{gu2019hplflownet,liu2019flownet3d,wang2021hierarchical,wu2020pointpwc}, but their accuracy is limited due to their local flow embedding design, which can only capture local matches. In contrast to these approaches, other architectures show more robust results by capturing global correlations and iteratively refining the correlation locally, based on Gated Recurrent Units (GRUs) to estimate highly accurate scene flow  \cite{gu2022rcp,kittenplon2021flowstep3d,wei2021pv}. However, iterative refinement of these approaches requires repeated flow embedding, which is computationally expensive and less efficient than hierarchical approaches.
In contrast to all the above approaches, we follow FLOT \cite{puy20flot} by using optimal transport to capture soft correspondences based on the Sinkhorn algorithm \cite{cuturi2013sinkhorn,peyre2019computational}. This algorithm is nonparametric and robust when strong features are extracted from the point cloud. To this end, rather than relying solely on point cloud features, we incorporate RGB features to enhance them, as RGB information is richer, particularly in geometrically homogeneous regions.

\subsection{Fusion-Based Methods for Scene Flow Estimation}
Cameras and LiDAR provide complementary information that benefits tasks like scene flow estimation. 
Many methods combine these two modalities to address their individual limitations and improve overall performance. 
For example, LiDAR-Flow \cite{battrawy2019lidar} combines information from stereo images and LiDAR to produce more accurate results, while DeepLiDARFlow \cite{rishav2020deeplidarflow} is a learning-based method that fuses monocular images and LiDAR measurements to produce a robust and dense representation of scene flow. All of the above methods work primarily in the image domain to extract pixel-wise matches and decode them into scene flow. They are also prone to feature propagation errors and are computationally expensive. Under the 3D rigidity assumption, RAFT-3D \cite{teed2021raft} uses dense RGB-D to estimate scene flow based on iterative refinements of the optical flow along with 3D homogeneous transformation matrices applied to each object in the scene at the expense of efficiency. 

More recently, coarse-to-fine fusion-based approaches such as CamLiFlow \cite{liu2021camliflow} and DELFlow \cite{peng2023delflow} introduced multi-stage bidirectional fusion between RGB and point cloud branches across feature extraction, cost volume construction, and flow estimation. These methods estimate dense optical flow (2D) from the RGB branch and sparse scene flow (3D) from the point cloud branch, with DELFlow supporting higher-density point clouds than CamLiFlow. 
However, the pipelines of both require more parameters and more expensive computations, and they require high-performance GPUs with large memory capacity. Compared to these methods, our \name{} works on a single-stage fusion, avoiding multiple cost volumes and multiple flow estimators. As a result, our pipeline has fewer parameters, a smaller memory footprint and does not require high-performance GPUs.
On the other hand, some approaches use monocular RGB intensities to augment point cloud features and operate solely in the 3D domain for correlation estimation \cite{wang2021festa,wang2020flownet3d++}. These methods typically concatenate RGB values directly with the corresponding 3D coordinates (\ie, xyz). However, their accuracy remains limited because they do not fully exploit rich RGB features. In contrast, \name{} extracts deep RGB representations to enhance point cloud features while preserving the efficiency of our design, making it easier for the Sinkhorn solver to compute robust correlations.

\section{\name{} Architecture } \label{SF3D-RGB Architecture}
\begin{figure*}[t]
	\begin{center}
		\includegraphics[width=1.0\linewidth]{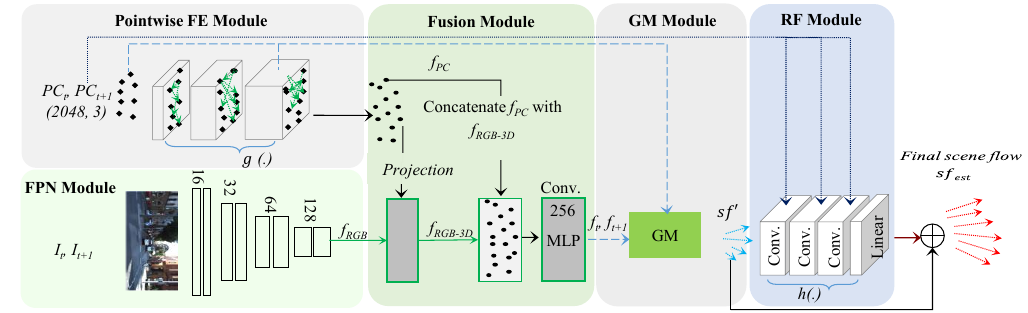}
		\caption{\name{} consists of a Pointwise Feature Extraction (FE) module, a Feature Pyramid Network (FPN), a Fusion Module (FM), a Graph Matching Module (GM), and a Refinement Module (RF). We denote the feature extraction convolutions by $g(\cdot)$.}
		\label{fig:SF3D-RGB_pipeline}
	\end{center}
\end{figure*}
Our \name{} architecture works with consecutive RGB images $(I_{t}, I_{t+1})$ both $\in \mathbb{R}^{H\times W\times 3}$  and the corresponding LiDAR scans $\big(PC_{t} = \{\mathbf{p}_{i} \in \mathbb{R}^{3}\}^N_{i=1},\ PC_{t+1} = \{\mathbf{p}_{j} \in \mathbb{R}^{3}\}^N_{j=1}\big)$, represented as 3D coordinates (\ie, xyz). Our method fuses the coarsest-level features of $(I_{t}, I_{t+1})$ and $(PC_{t}, PC_{t+1})$ to produce a 3D prediction of the scene flow. 
We assume that both the LiDAR and the monocular camera are well-calibrated.
Our architecture consists of five modules (\cf\cref{fig:SF3D-RGB_pipeline}): RGB Feature Pyramid Network (FPN), Pointwise Feature Extraction Module (FE), Fusion Module (FM), Graph Matching Module (GM) and Refinement Flow Module (RF). Each module of the architecture is described in detail below:

\subsection{Feature Pyramid Network Module}
The feature pyramid network (FPN) module extracts multiscale features with strong semantics and localization from RGB images $(I_{t}, I_{t+1})$ using the FPN pipeline \cite{lin2017feature}. The FPN has four levels of doubled stride convolutions (16, 32, 64, and 128) with a negative slope of 0.1, LeakyReLU activation, and an instance normalization layer. It encodes monocular RGB images into features with reduced resolution at each downscaled level of the pyramid.

\subsection{Feature Extraction Module of Point Clouds}
\begin{figure*}[t]
	\begin{center}
		\includegraphics[width=1.0\linewidth]{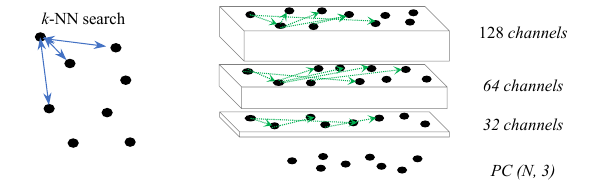}
		\caption{Pointwise Feature Extraction Module based on graph convolution.}
		\label{fig:pointwise_featuremap}
	\end{center}
\end{figure*}

The pointwise feature extraction (FE) module is inspired by \cite{qi2017pointnet} and uses graph convolution to extract unordered raw point cloud features from $(PC_{t}, PC_{t+1})$ (\ie, no intermediate representation).
Instead of hierarchical sampling, we operate at the full input resolution by passing the point cloud (\ie, $PC \in \mathbb{R}^{N \times 3}$) through a series of Multi-Layer Perceptrons (MLPs) to extract per-point features $\mathbf{f}_{\mathbf{p}_{i}}^{(l)} \in \mathbb{R}^{c}$, where $l$ denotes the layer and $c$ is the feature channel dimension. The channel dimension changes from one layer to the next, as shown in \cref{fig:pointwise_featuremap}.
For each centroid, neighboring points are selected using a fixed k-nearest neighbors (\textit{k}-NN) scheme with $knn=32$.
The indices of the neighboring points are denoted by $\mathcal{N}(\mathbf{p}_{i})$.

For each neighbor $\mathbf{p}_{j}$ of $\mathbf{p}_{i}$, we construct an edge feature by concatenating the neighbor’s feature with the relative spatial offset:
\begin{equation}
\mathbf{e}_{ij}^{(l)} = \left[\left(\mathbf{f}_{\mathbf{p}_{j}}^{(l)}\right)^{\top},\ \left(\mathbf{p}_{j} - \mathbf{p}_{i}\right)^{\top}\right]^{\top} \in \mathbb{R}^{c}
\end{equation}
where $j \in \mathcal{N}(\mathbf{p}_{i})$, and $(\cdot)^{\top}$ represents transposition. 
The local regions are then encoded into feature vectors using three convolutional layers (32, 64, and 128) as shown in \cref{fig:pointwise_featuremap}. The features are passed through an MLP $g: \mathbb{R}^{c} \to \mathbb{R}^{c''}$ with fully connected layers and LeakyReLU activation with a negative slope of 0.1, repeated three times, and finally maximum pooled. This is done for each point $\mathbf{p}_{i}$ in the point cloud.

\begin{equation}
\mathbf{f}_{\mathbf{p}_{i}}^{(l+1)} = \max_{j \in \mathcal{N}(\mathbf{p}_{i})} g\!\left(\mathbf{e}_{ij}^{(l)}\right) \in \mathbb{R}^{c''}
\end{equation}
The resulting features $\mathbf{f}_{\mathbf{p}_{i}}^{(l+1)}$ serve as input to the next layer. Stacking several such layers yields the final point cloud features ($f_{PC}$).
 
\subsection{Fusion Module}
We fuse the most informative cues from RGB and LiDAR data in our fusion module (FM), adopting a late fusion strategy \cite{eitel2015multimodal,eldesokey2019confidence,rishav2020deeplidarflow}, where the coarsest RGB features ($f_{RGB}$) are concatenated with the point cloud features ($f_{PC}$). 
To obtain the corresponding RGB features, each 3D point is projected onto the image plane using the camera intrinsics. The concatenated features ($f_{RGB-3D}$) are then passed through a 256-channel MLP, producing fused features. Since this process is applied to consecutive LiDAR and RGB frames, we denote the fused features at timestamp $t$ and $t+1$ by $f_{t}$ and $f_{t+1}$, respectively.
These fused features are subsequently used by the graph matching module to compute the initial scene flow ($sf'$). In \cref{Ablation Study}, early and late fusions are compared to choose the best option.

\subsection{Optimal Transport for Graph Matching} \label{Optimal Transport for Graph Matching}
\begin{algorithm}[t]
\caption{Optimal Transport Module}
\label{alg:ot}
\begin{algorithmic}[1]
\Require Transport cost matrix $\mathbf{C}$; entropy parameter $\varepsilon \ge 0$; marginals $\mu_{s}, \mu_{t}$; iteration count $k \leftarrow 1$
\Ensure Optimal transport matrix $\mathbf{T}^{\ast}$
\State $\mathbf{K} \gets \exp(-\mathbf{C}/\varepsilon)$
\State $\mathbf{a} \gets \mathbf{1}_{N},\quad \mathbf{b} \gets \mathbf{1}_{N}$
\For{$\ell = 1$ to $k$}
    \State $\mathbf{b} \gets \mu_{t} \oslash (\mathbf{K}^{\top}\mathbf{a})$
    \State $\mathbf{a} \gets \mu_{s} \oslash (\mathbf{K}\mathbf{b})$
\EndFor
\State $\mathbf{T}^{\ast} \gets \operatorname{Diag}(\mathbf{a})\,\mathbf{K}\,\operatorname{Diag}(\mathbf{b})$
\State \Return $\mathbf{T}^{\ast}$
\Statex \textbf{Note:} $\oslash$ denotes element-wise division; multiplications are element-wise where applicable.
\end{algorithmic}
\end{algorithm}
The goal of the graph matching module (GM) is to find the minimum-cost transport (\ie, optimal transport) from a source distribution $\mu_{s}$ to a target distribution $\mu_{t}$, where $\mu_{s}, \mu_{t}$ are probability measures on $\mathbb{R}^{N}$. 
We follow \cite{puy20flot}: features $f_{t}$ and $f_{t+1}$ from the fusion module form the transport cost matrix $\mathbf{C}$, and the displacement cost from point clouds supports finding the transport plan $\mathbf{T}^{\ast}$ via
\begin{equation}
\label{eq:ot}
\mathbf{T}^{\ast} = \arg\min_{\mathbf{T} \in \mathbb{R}_{+}^{N \times N}} \sum_{i,j} C_{ij} T_{ij}
\end{equation}
subject to
\begin{equation}
\label{eq:ot_constraints}
\begin{aligned}
\mathbf{T}\,\mathbf{1}_{N} &= \mu_{s},\\
\mathbf{T}^{\top}\mathbf{1}_{N} &= \mu_{t},
\end{aligned}
\end{equation}
where $\mathbf{1}_{N}$ is the all-ones vector. Here $\mathbf{T}^{\ast}$ is the optimal transport plan; $T_{ij}$ denotes the mass transported from source $i$ to target $j$ to minimize the cost, and $C_{ij} \ge 0$ is the displacement cost.
\Cref{eq:ot_constraints} applies two constraints where the first constraint ensures that each $f_{t}$ is fully distributed over $f_{t+1}$, while the second constraint ensures that each $f_{t+1}$ receives a mass $\mu_{s}$ from $f_{t}$. 

We compute the displacement cost in the same manner as FLOT \cite{puy20flot}, but using our fused features by measuring cosine distances in feature space. The resulting cost matrix entries are given by:
\begin{equation}
\label{eq:cost}
C_{ij} = \left(1 - \frac{\mathbf{f}_{t,i}^{\top} \mathbf{f}_{t+1,j}}{\|\mathbf{f}_{t,i}\|_{2}\,\|\mathbf{f}_{t+1,j}\|_{2}}\right)\,\mathbb{1}\!\left(\|\mathbf{p}_{t,i} - \mathbf{p}_{t+1,j}\|_{2} \le d_{\max}\right),
\end{equation}

\begin{equation}
\mathbb{1}\!\left(\|\mathbf{p}_{t,i} - \mathbf{p}_{t+1,j}\|_{2} \le d_{\max}\right) =
\begin{cases}
1, & \|\mathbf{p}_{t,i} - \mathbf{p}_{t+1,j}\|_{2} \le d_{\max},\\
0, & \text{otherwise.}
\end{cases}
\end{equation}

where $\mathbb{1}(\cdot)$ is the indicator function that removes correspondences with displacement greater than $d_{\max}=10$m.

In an ideal case without viewpoint changes or occlusions, a point in the source frame maps exactly to its corresponding point in the target frame under the ground-truth scene flow \cite{puy20flot}. In practice, occlusions violate this assumption, causing the mass-preservation constraints in \cref{eq:ot,eq:ot_constraints} to fail. Following \cite{puy20flot}, we introduce a Kullback–Leibler (KL) divergence term to relax these constraints when mass is not preserved. The formulation of \cref{eq:ot} then becomes:
\begin{equation}
\label{eq:entropic_ot}
\mathbf{T}^{\ast} = \arg\min_{\mathbf{T} \in \mathbb{R}_{+}^{N \times N}} \sum_{i,j} C_{ij} T_{ij} + \varepsilon\,H(\mathbf{T}) + \lambda\Big(\mathrm{KL}(\mathbf{T}\mathbf{1}_{N} \,\|\, \mu_{s}) + \mathrm{KL}(\mathbf{T}^{\top}\mathbf{1}_{N} \,\|\, \mu_{t})\Big),
\end{equation}
where the entropy term is $H(\mathbf{T}) = \sum_{i,j} T_{ij}(\log T_{ij}-1)$, $\mathrm{KL}(\cdot\|\cdot)$ is the Kullback–Leibler divergence, and $\lambda,\varepsilon \ge 0$ are learnable parameters optimized with the network. 
A smaller ($\varepsilon$) yields a sparser transport plan, and ($\lambda$) controls how much the marginals deviate from uniformity, allowing mass variation. We use the Sinkhorn algorithm \cite{cuturi2013sinkhorn,peyre2019computational} to efficiently estimate the transport map with these regularizers.
The complete algorithm is described in \cref{alg:ot}.

The entropic regularization strength is governed by $\lambda/(\lambda + \varepsilon)$: as $\lambda \to \infty$ enforces stricter mass preservation, and as $\lambda \to 0$ (or with occlusions) the regularization weakens. 
We obtain the initial scene flow $\mathbf{sf}'_{i}$ using the soft assignment matrix computed from $\mathbf{T}^{\ast}$, following \cite{puy20flot}, but with fused features instead of only point cloud features: 
\begin{equation}
\label{eq:sf_initial}
\mathbf{sf}'_{i} = \left(\sum_{j=1}^{N} T_{ij}\,\mathbf{p}_{t+1,j}\right)\big/ \left(\sum_{j=1}^{N} T_{ij}\right) - \mathbf{p}_{t,i}
\end{equation}
\subsection{Refinement Flow Module and Scene Flow Estimation}
The architecture of the Refinement Flow (RF) module $h(\cdot)$ is the same as $g(\cdot)$ for point cloud feature extraction, but it includes an extra linear layer with an MLP size of 3 to learn correlations between inputs and outputs \cite{puy20flot}. The RF module uses the correspondences to refine the initial scene flow estimate produced by the optimal transport $\mathbf{T}^{\ast}$. Since $PC_{t}$ and $PC_{t+1}$ provide metric distances and correspondences, some points in $PC_{t}$ may still lack exact matches in $PC_{t+1}$. The refinement module applies a residual network to adjust the flow accordingly as clarified in \cref{eq:refine}.

\begin{equation}
\label{eq:refine}
\mathbf{sf_{i,\text{est}}} = \mathbf{sf_i}' + MLP(h(\mathbf{sf_i}'))
\end{equation}
where $h: \mathbb{R}^{3} \to \mathbb{R}^{c}$ takes the initial estimated scene flow $\mathbf{sf_i}'$ as input. Its output is then passed through an MLP to produce the final scene flow estimate $\mathbf{sf_{i,est}}$. We use $i$ to denote a single point.

\section{Experiments} \label{Experiments}
Our method is validated through several experiments, including comparison with the state-of-the-art, verification of our design decisions, and comparison with other fusion-based solutions to show performance and efficiency.  
\subsection{Datasets and Evaluation Metrics} \label{Datasets and Evaluation Metrics}

\setcounter{footnote}{0}

As in related work, we train our network with full supervision on FlyingThings3D (FT3D) \cite{mayer2016large} and evaluate on FT3D and KITTI Scene Flow \cite{menze2015object}. Below, we outline the datasets, preprocessing steps, and evaluation metrics used in our experiments. 

\textbf{FlyingThings3D (FT3D) \cite{mayer2016large}} is a large-scale synthetic dataset for scene flow estimation. Following \cite{gu2019hplflownet}, we adopt the curated split that removes extremely difficult examples and discard points beyond 35 meters or with occluded disparity/optical flow. Points are independently sampled per frame without enforcing correspondences. FT3D provides 19,640 training examples and 3,824 test examples; from each scan we draw $N=2048$ unpaired points. For training and evaluation, we use left-view RGB images (\ie, monocular) and convert disparity to 3D to form the point clouds.

\textbf{stereoKITTI \cite{menze2015object}} is a small real-world dataset with optical flow and disparity annotations for 142 labeled scenes from the KITTI raw data. Disparity maps are captured by LiDAR and densified on dynamic objects. The disparity of the second frame at time $t{+}1$ is warped into the reference view at time $t$, which induces strong correlations between consecutive point clouds. To decorrelate the second frame and generate the point cloud $PC_{t+1}$ at timestamp $t{+}1$, we apply the DeepLiDARFlow preprocessing \cite{rishav2020deeplidarflow}\footnote{\url{https://github.com/dfki-av/DeepLiDARFlow}.} and refer to the resulting dataset as KITTId. 
We split the 200 examples into 180 for fine-tuning and 20 for validation, remove 58 scenes with incomplete or noisy annotations, and retain 142 scenes for evaluation. For comparisons without fine-tuning, we include all scenes. For fair LiDAR-based comparisons, we exclude ground points during training and evaluation. We independently sample $N=2048$ non-corresponding random points from each point cloud scene. 
For training and evaluation, we use left-view RGB images (\ie, monocular) and convert disparity to 3D to form the point clouds. 

\textbf{lidarKITTI~\cite{geiger2012we}} differs from stereoKITTI in that consecutive LiDAR scans are not perfectly aligned and naturally contain occlusions between frames. Scenes are represented as raw point clouds, preserving the irregular patterns of real LiDAR without densification or extra preprocessing. Although it covers the same 142 annotated stereo pairs, scene flow ground truth is obtained by projecting LiDAR points into the image domain and associating them with stereoKITTI pixels. We evaluate models on the lidarKITTI test split (14 frames) after fine-tuning on KITTId. Without fine-tuning, we use all 142 scenes; with fine-tuning, we split the scenes into 128 for training and 14 for validation. As with KITTId, we exclude ground points during training and evaluation and independently draw a non-corresponding random sample of $N=2048$ points from each scan.

We use five standard metrics, consistent with \cite{gu2019hplflownet,liu2021camliflow,liu2019flownet3d,puy20flot,teed2021raft,wu2020pointpwc}, averaged over all scenes and points and defined as follows:
\begin{itemize}[noitemsep,nolistsep,topsep=1pt,leftmargin=*,label=\textbullet]
	\item {\textit{EPE3D [m]}}: The 3D end-point error computed in meters as $\|\mathbf{sf}_{\text{est}} - \mathbf{sf}_{\text{gt}}\|_{2}$. 
	\item {\textit{Acc3DS [\%]}}: The strict 3D accuracy which is the ratio of points whose \textit{EPE3D} $< 0.05~m$ \textbf{or} relative error $< 5\%$. 
	\item {\textit{Acc3DR [\%]}}: The relaxed 3D accuracy which is the ratio of points whose \textit{EPE3D} $< 0.1~m$ \textbf{or} relative error $< 10\%$. 
	\item {\textit{Out3D [\%]}}: The ratio of outliers whose \textit{EPE3D} $> 0.3~m$ \textbf{or} relative error $> 10\%$. 
    \item {\textit{EPE2D [pix]}}: The 2D end-point error in pixels computed after projecting estimated and ground truth scene flow onto the image plane.
\end{itemize}

\subsection{Implementation and Training} \label{Implementation and Training}
For training at a specific resolution, the preprocessed consecutive frames are randomly sub-sampled to $N=2048$ points and reshuffled. 
The training loss uses the $\ell_{1}$ norm of the masked flow difference:
\begin{equation}
\min_{\theta} \frac{1}{3L} \sum_{l = 1}^{L} \left\| \mathbf{M}^{(l)} \big( \mathbf{sf}_{\text{est}}^{(l)} - \mathbf{sf}_{\text{gt}}^{(l)} \big) \right\|_{1},
\end{equation}
where $\mathbf{sf}_{\text{gt}}$ is the ground-truth scene flow, $\mathbf{sf}_{\text{est}}$ is the predicted flow, and $\mathbf{M}^{(l)}$ is a binary mask that removes points with occluded flow.

We train our architecture on FT3D \cite{mayer2016large} with full supervision and a batch size of 4 using Adam \cite{kinga2015method}. The learning rate begins at 0.001, is reduced to 0.0001 after 340 epochs, and training continues for an additional 60 epochs. For fine-tuning on KITTId and lidarKITTI, we train for 200 epochs: a fixed learning rate of 0.001 for the first 100 epochs, then 0.0001 for the remaining 100. 
Following FLOT \cite{puy20flot}, we learn $(\theta, \varepsilon, \lambda)$; to keep $\varepsilon$ and $\lambda$ positive, we optimize their logarithms and apply an exponential with a 0.03 offset to avoid numerical issues. 

\subsection{Comparison with Related Work}  \label{Comparison to Related Work}
\begin{table}[t]
\caption{Quantitative comparison on the FlyingThings3D (FT3D) test set. $\downarrow$ indicates lower is better. We categorize the table into three groups based on the scene flow estimation domain (\ie, 2D, 2D/3D, or 3D). Model size is evaluated using the number of parameters (P). Runtime (R) is analyzed across different hardware devices, and the (TFLOPS) column gives the tera-floating-point-operations per second; higher values indicate more powerful GPUs. Best values in each category, with and without fine-tuning, are highlighted in bold.}
\label{tab:ft3d}
\centering
\setlength{\tabcolsep}{4pt}
\resizebox{\textwidth}{!}{%
\begin{tabular}{|l!{\vrule width 1pt}l|c|c|cc|cccc|}
\hline
\multirow{3}{*}{\textbf{Method}} & \multirow{3}{*}{\textbf{Modality}} & \multirow{2}{*}{\textbf{Evaluated}} & \multirow{3}{*}{\textbf{Domain}} & \multicolumn{2}{c|}{\textbf{FT3D}} & \multicolumn{4}{c|}{\textbf{Model size and runtime}} \\ \cline{5-10}
 &  &  &  & \textbf{EPE3D} & \textbf{EPE2D} & \textbf{P} & \textbf{R} & \textbf{TFLOPS} & \textbf{Device} \\
 &  & \textbf{Points} &  & [m] $\downarrow$ & [px] $\downarrow$ & [M] $\downarrow$ & [ms] $\downarrow$ & & \\ \hline\hline
PWOC-3D \cite{saxena2019pwoc} & Stereo only & all   & 2D    & --    & 6.97 & 8.05  & 130   & 11.34 & GTX1080Ti \\ \hline
LiDAR-Flow \cite{battrawy2019lidar} & Stereo+LiDAR & 5000 & 2D   & --    &29.97 & --    &65900 & - & CPU \\ \hline
DeepLiDARFlow \cite{rishav2020deeplidarflow} & RGB+LiDAR & 2048 & 2D   & --    & 6.41 & 8.29  & 82   & 13.45 & RTX2080Ti \\ \hline
RAFT-3D \cite{teed2021raft} & RGB-D & all & 2D & \textbf{0.094} & \textbf{2.37} &45.00 & 386 & 11.34 & GTX1080Ti \\ \hline\hline
CamLiFlow \cite{liu2021camliflow} & RGB+LiDAR & 2048 & 2D/3D & 0.097 & 4.89 & 7.70 & 133 & 10.07 & RTX2080 \\ \hline
CamLiFlow \cite{liu2021camliflow} & RGB+LiDAR & 8192 & 2D/3D &\textbf{0.061} & \textbf{2.20} & 7.70 & 155 & 10.07 & RTX2080 \\ \hline
DELFlow \cite{peng2023delflow} & RGB+LiDAR & 8192 & 2D/3D &\textbf{0.061} & 2.23 &20.10 &  85 & 38.77 & RTX A6000 \\ \hline\hline
FlowNet3D \cite{liu2019flownet3d} & LiDAR only & 2048 & 3D & 0.134 & -- & 1.20 & 64 & 11.34 & GTX1080Ti \\ \hline
PointPWC-Net \cite{wu2020pointpwc} & LiDAR only & 2048 & 3D & 0.121 & -- & 5.30 & 59 & 11.34 & GTX1080Ti \\ \hline
FLOT \cite{puy20flot} & LiDAR only & 2048 & 3D & 0.156 & - & \textbf{0.12} & \textbf{26} & 13.45 & RTX2080Ti \\ \hline
FlowNet3D++ \cite{wang2020flownet3d++} & RGB+LiDAR & 8192 & 3D & 0.137 & -- & -- & 64 & 11.34 & GTX1080Ti \\ \hline
FESTA \cite{wang2021festa} & RGB+LiDAR & 8192 & 3D & 0.113 & -- &16.10 & 68 & 11.34 & GTX1080Ti \\ \hline
\textbf{\name{} (ours)} & \textbf{RGB+LiDAR} & \textbf{2048} & \textbf{3D} & \textbf{0.102} & \textbf{5.03} & 0.48 & 39 & 13.45 & RTX2080Ti \\ \hline
\end{tabular}%
}
\end{table}
In terms of accuracy and efficiency, we compare \name{} with state-of-the-art methods, including those using different input modalities (\cref{tab:ft3d}). 
We categorize the table into three groups based on the scene flow estimation domain (\ie, 2D, 2D/3D, or 3D).
Although our method fuses 2D image features with 3D point features, we still consider it a 3D approach, since the final scene flow prediction is performed entirely in 3D after the fusion module.
\begin{table}[t]
\caption{Metrics and comparison on KITTId and lidarKITTI. FT denotes fine-tuning. Best values in each category, with and without fine-tuning, are highlighted in bold.}
\label{tab:kitti_results}
\centering
\setlength{\tabcolsep}{3pt}
\resizebox{\textwidth}{!}{%
\begin{tabular}{|l!{\vrule width 1pt}l|c|ccc|ccc|}
\hline
\multirow{3}{*}{\textbf{Method}} & \multirow{3}{*}{\textbf{Modality}} & \multirow{3}{*}{\textbf{FT}} &
\multicolumn{3}{c|}{\textbf{KITTId}} &
\multicolumn{3}{c|}{\textbf{lidarKITTI}} \\ \cline{4-9}
 &  &  & \textbf{EPE3D} & \textbf{Acc3DR} & \textbf{Acc3DS} & \textbf{EPE2D} & \textbf{EPE3D} & \textbf{Acc3DR} \\
  &  &  & \textbf{[m]} $\downarrow$ & \textbf{[\%]} $\uparrow$ & \textbf{[\%]} $\uparrow$ & \textbf{[px]} $\downarrow$ & \textbf{[m]} $\downarrow$ & \textbf{[\%]} $\uparrow$ \\ \hline\hline
FLOT \cite{puy20flot} & LiDAR only & $\times$ & 0.311 & 48.5 & 25.4 & 18.19 & 0.501 & 33.7 \\ \hline
FlowStep3D \cite{kittenplon2021flowstep3d} & LiDAR only & $\times$ & 0.303 & 57.8 & 38.2 & -- & 0.479 & 20.4 \\ \hline
\textbf{\name{} (ours)} & \textbf{RGB+LiDAR} & $\times$ & \textbf{0.227} & \textbf{63.1} & \textbf{40.1} & \textbf{14.46} & \textbf{0.395} & \textbf{49.8} \\ \hline\hline
FLOT \cite{puy20flot} & LiDAR only & $\checkmark$ & 0.101 & 80.2 & 57.8 & 5.48 & 0.289 & 66.2 \\ \hline
DeepLiDARFlow \cite{rishav2020deeplidarflow} & RGB+LiDAR & $\checkmark$ & 0.209 & -- & -- & 3.18 & 0.349 & -- \\ \hline
CamLiFlow \cite{liu2021camliflow} & RGB+LiDAR & $\checkmark$ & \textbf{0.067} & \textbf{91.0} & \textbf{75.4} & 4.52 & \textbf{0.128} & \textbf{80.5} \\ \hline
\textbf{\name{} (ours)} & RGB+LiDAR & $\checkmark$ & 0.078 & 87.5 & 71.3 & \textbf{4.46} & 0.265 & 71.9 \\ \hline
\end{tabular}%
}
\end{table}

\textbf{Comparison on FT3D dataset:}
In terms of accuracy, by fusing monocular RGB and LiDAR data, our method achieves significant improvement compared to LiDAR-only approaches \cite{liu2019flownet3d,wu2020pointpwc,puy20flot} on FT3D. Notably, it also outperforms FLOT \cite{puy20flot}, the LiDAR-only baseline that employs a similar transport mechanism. Incorporating RGB features yields a substantial improvement in performance.
We also outperform early fusion approaches like \cite{wang2020flownet3d++,wang2021festa}, whose architectures operate fully in 3D. 
Compared to methods that rely on stereo systems, such as PWOC-3D \cite{saxena2019pwoc}, and those that fuse LiDAR with stereo \cite{battrawy2019lidar}, our approach achieves substantially lower EPE2D.
Furthermore, our method also outperforms DeepLiDARFlow \cite{rishav2020deeplidarflow}.
In terms of model size, apart from FLOT \cite{puy20flot}, \name{} uses fewer trainable parameters.
Since we report runtime across different hardware devices, we use tera-floating-point-operations per second (TFLOPS) from the hardware specifications to indicate GPU capability, where higher TFLOPS reflects a more powerful GPU. 
Notably, we report results for our approach, FLOT \cite{puy20flot}, and DeepLiDARFlow \cite{rishav2020deeplidarflow} on RTX2080Ti. 
On this device, our method runs faster than DeepLiDARFlow.
Our approach also reports lower runtime than 3D-domain methods \cite{liu2019flownet3d,wu2020pointpwc,wang2021festa,wang2020flownet3d++}, even though those runtimes are measured on a GTX1080Ti with about 0.85$\times$ the TFLOPS of our hardware (11.34 vs. 13.45). 
Given the small difference between the two GPUs, we expect our method will still run faster on the same hardware (\ie, GTX1080Ti).
Compared to the rigidity-based method RAFT-3D \cite{teed2021raft}, our approach uses fewer parameters, is more efficient, and operates directly at the point level rather than relying on rigidity assumptions, while achieving comparable EPE3D and EPE2D results.
We compare our single-stage fusion to CamLiFlow’s multi-stage fusion \cite{liu2021camliflow} (using 2048 points) and obtain comparable results while requiring fewer parameters. 
However, with denser point inputs (\ie, 8192), the multi-stage fusion methods DELFlow \cite{peng2023delflow} and CamLiFlow \cite{liu2021camliflow} achieve higher accuracy than our approach, but at the cost of efficiency. 
Among all state-of-the-art methods, and except for FLOT \cite{puy20flot}, \name{} requires fewer trainable parameters and provides faster, more efficient inference, offering a strong trade-off between accuracy and efficiency.

\textbf{Comparison on KITTId and lidarKITTI datasets:}
Our \name{} is also evaluated on KITTId and lidarKITTI (see \cref{tab:kitti_results}). 

Using the 142 annotated scenes described in \cref{Datasets and Evaluation Metrics}, we evaluate \name{} against LiDAR-only methods \cite{kittenplon2021flowstep3d,puy20flot} without any fine-tuning and observe notably higher accuracy. In particular, compared to the LiDAR-only baseline FLOT \cite{puy20flot}, our method shows a substantial improvement, demonstrating that fusing RGB features significantly boosts overall accuracy on both datasets KITTId and lidarKITTI.

We also fine-tune our method and FLOT \cite{puy20flot} following the protocol of \cite{liu2021camliflow,rishav2020deeplidarflow}, using the same annotated frames as DeepLiDARFlow \cite{rishav2020deeplidarflow}. All methods \cite{puy20flot,liu2021camliflow,rishav2020deeplidarflow} are fine-tuned on the same split and evaluated on the remaining frames (20 for KITTId and 14 for lidarKITTI). The road surface is excluded for all methods during both fine-tuning and evaluation. The results show that our method consistently outperforms the LiDAR-only baseline FLOT across all metrics on both datasets, with and without fine-tuning. 
With fine-tuning, our method with the fusion of monocular camera and LiDAR outperforms FLOT \cite{puy20flot} on both KITTId and lidarKITTI. 
However, CamLiFlow \cite{liu2021camliflow} with 2048 points remains more accurate but is less efficient, as shown in \cref{tab:ft3d}. 
\begin{figure*}[t]
	\begin{center}
		\includegraphics[width=1.0\linewidth]{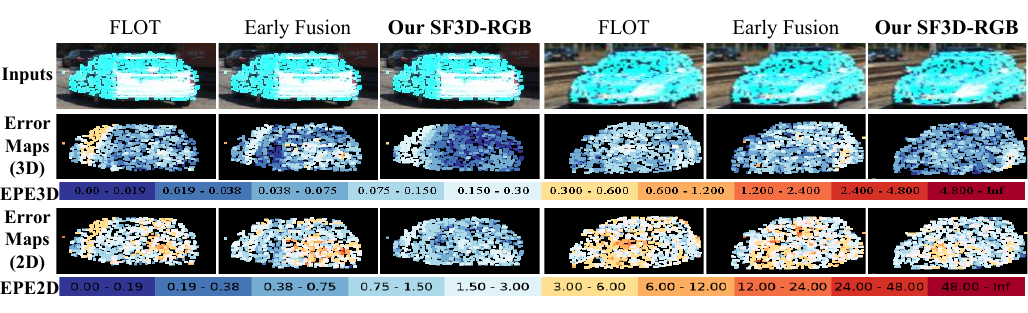}
		\caption{Qualitative scene flow results of \name{} on the KITTId dataset, compared with LiDAR-only and early fusion baselines. LiDAR points are overlaid on the images to improve visualization clarity of the input scenes. Note that FLOT \cite{puy20flot} uses only LiDAR as input, whereas both the early fusion and our \name{} use LiDAR combined with a monocular camera as inputs. In the error maps, dark blue indicates lower error, dark red indicates higher error, and black regions indicate unavailable LiDAR points due to removal of ground points.}
		\label{fig:KITTId_images}
	\end{center}
\end{figure*}

\begin{figure*}[t]
	\begin{center}
		\includegraphics[width=1.0\linewidth]{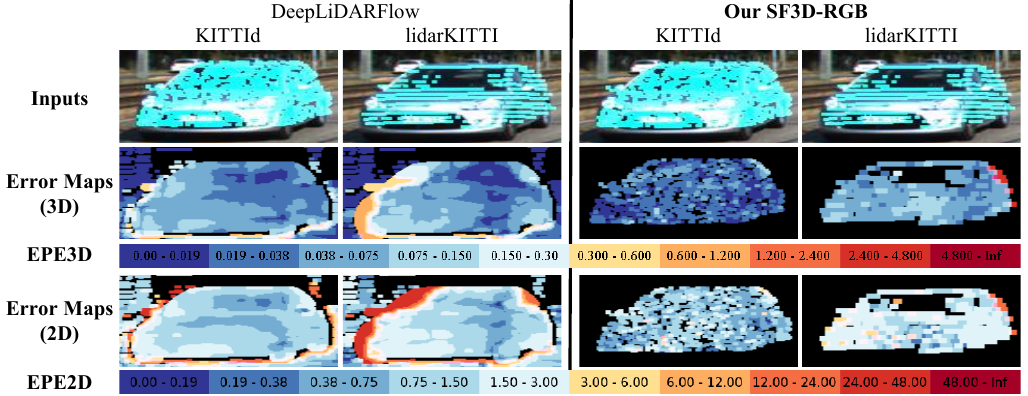}
		\caption{Qualitative comparison of \name{} and DeepLiDARFlow \cite{rishav2020deeplidarflow} on the KITTId and lidarKITTI test datasets. Our \name{} results show strong scene flow accuracy across both datasets compared to DeepLiDARFlow. LiDAR points are overlaid on the images to improve visualization clarity of the input scenes. Note that DeepLiDARFlow and our \name{} use LiDAR combined with a monocular camera as inputs. In the error maps, dark blue indicates lower error, dark red indicates higher error, and black regions indicate unavailable LiDAR points due to removal of ground points. Note that DeepLiDARFlow does not exclude ground points in its scene flow estimation.}
		\label{fig:lidarKITTI_images}
	\end{center}
\end{figure*}

% \subsection{Qualitative Comparison}  \label{Qualitative Results}

We provide a visual comparison of the scene flow estimates in \cref{fig:KITTId_images,fig:lidarKITTI_images}. \Cref{fig:KITTId_images} presents the results on KITTId, showing how \name{} compares to FLOT (LiDAR-only) and early fusion. The error maps of our fusion method consistently exhibit lower EPE3D and EPE2D, demonstrating the clear effect of our fusion compared to FLOT \cite{puy20flot}. \Cref{fig:lidarKITTI_images} compares \name{} with DeepLiDARFlow \cite{rishav2020deeplidarflow} using the KITTId and lidarKITTI versions. It shows that our fusion in \name{} has better accuracy than DeepLiDARFlow, which considers LiDAR as a depth map representation. 

% While DeepLiDARFlow estimates the scene flow in 2D representation and in a dense way, we need to include the road surface in the evaluation of DeepLiDARFlow to achieve a high accuracy.

\subsection{Ablation Study} \label{Ablation Study}
We conduct several experiments to validate our design choices. FLOT \cite{puy20flot} is trained with the same number of points, and we apply early fusion of point cloud and RGB intensities for a fair comparison. \name{} is then trained on FlyingThings3D \cite{mayer2016large}, and the results are summarized in \cref{tab:ablation_ft3d}. The findings show that early fusion already improves accuracy over the LiDAR-only baseline, while our final fusion strategy in \name{} yields a substantial further improvement across all evaluation metrics. 
With fine-tuning on the KITTId dataset, we further validate our design choices by analyzing the impact of different components on accuracy, as shown in \cref{tab:arch_kitti}. Based on the EPE3D and OUT3D metrics, fusing RGB and point cloud features at the coarsest level using a single MLP yields the best performance. This setup outperforms using LiDAR alone as well as early fusion. We also observe that using a single MLP in the fusion module (FM) (\cf\cref{fig:SF3D-RGB_pipeline}) achieves better results than using two MLPs (\ie, {1$\times$MLP} vs. {2$\times$MLP}).

Furthermore, we investigate in \cref{tab:ablation_ft3d,tab:arch_kitti} the trainable entropic regularization ratio $\lambda/(\lambda + \varepsilon)$ to identify the setting that yields the best performance.

\begin{table}[t]
\caption{Ablation study on FT3D (late fusion, number of Sinkhorn iterations $k$ fixed to 1). Best values are in bold.}
\label{tab:ablation_ft3d}
\centering
\setlength{\tabcolsep}{6pt}
\resizebox{\textwidth}{!}{%
\begin{tabular}{|l|c|cccc|}
\hline
\multirow{2}{*}{\textbf{Method}} & \multirow{2}{*}{$\mathbf{\dfrac{\lambda}{\lambda + \varepsilon}}$} & \textbf{EPE3D} & \textbf{OUT3D} & \textbf{Acc3DR} & \textbf{Acc3DS} \\
 &  & \textbf{[m]} $\downarrow$ & \textbf{[\%]} $\downarrow$ & \textbf{[\%]} $\uparrow$ & \textbf{[\%]} $\uparrow$ \\ \hline\hline
FLOT & 0.608 & 0.138 & 68.1 & 66.2 & 33.2 \\ \hline
Early Fusion & 0.549 & 0.129 & 64.2 & 70.3 & 36.1 \\ \hline
\textbf{\name{}} & \textbf{0.587} & \textbf{0.102} & \textbf{57.6} & \textbf{77.5} & \textbf{45.8} \\ \hline
\end{tabular}%
}
\end{table}

\begin{table}[t]
\caption{Verification of \name{} fusion on KITTId. Best values are in bold.}
\label{tab:arch_kitti}
\centering
\setlength{\tabcolsep}{6pt}
\resizebox{\textwidth}{!}{%
\begin{tabular}{|c|c|c|c|c|c|c|cc|}
\hline
\textbf{LiDAR} & \textbf{RGB} & \textbf{Early} & \textbf{Late} & \textbf{1$\times$MLP} & \textbf{2$\times$MLP} & \multirow{2}{*}{$\mathbf{\dfrac{\lambda}{\lambda + \varepsilon}}$} & \textbf{EPE3D} & \textbf{OUT3D} \\
\textbf{Modality} & \textbf{Modality} & \textbf{Fusion} & \textbf{Fusion} & \textbf{256} & \textbf{256} &  & \textbf{[m]} $\downarrow$ & \textbf{[\%]} $\downarrow$ \\ \hline\hline
\checkmark &  &  &  &  &  & 0.524 & 0.101 & 31.8 \\ \hline
\checkmark & \checkmark & \checkmark &  &  &  & 0.372 & 0.092 & 29.4 \\ \hline
\checkmark & \checkmark &  & \checkmark &  & \checkmark & 0.307 & 0.098 & 27.6 \\ \hline
\checkmark & \checkmark & & \checkmark & \checkmark &  & \textbf{0.443} & \textbf{0.078} & \textbf{25.5} \\ \hline
\end{tabular}%
}
\end{table}
\subsection{Limitations} \label{Limitations}
Our approach can work efficiently on low-density points (up to 4K) but struggles with high-density points due to the Sinkhorn-based soft correspondence. This can be alleviated by chunking high-density points into low-density partitions and regrouping the results. Another limitation is the need to omit the ground points (\ie, street surface) in real outdoor scenes.

\section{Conclusion} \label{Conclusion}
In this paper, we introduced \name{}, a fusion-based deep learning architecture that employs graph convolutional networks for 3D scene flow estimation. \name{} combines coarsest-level features from monocular RGB images and LiDAR point clouds to form robust fused representations. Our method demonstrates strong performance on real-world datasets with sparse LiDAR inputs (2048 points), surpassing LiDAR-only approaches in both accuracy and error metrics. It also delivers competitive accuracy with fewer parameters and is more efficient on low-power GPUs. For future work, we aim to extend our model to handle denser point clouds while maintaining the same level of efficiency.

\section*{Acknowledgments}
This work was partially funded by the Federal Ministry of Research, Technology, and Space Germany under the project COPPER (16IW24009).

\bibliographystyle{spmpsci}
\bibliography{bibliography}

\end{document}